# Deep Learning Models for Coral Bleaching Classification in Multi-Condition Underwater Image Datasets

## Julio Jerison E. Macrohon, PhD[*]

Computer Science Department, Hsinchu County American School, Zhubei City, Taiwan (ROC)

## Gordon Hung

Computer Science Department, Hsinchu County American School, Zhubei City, Taiwan (ROC)

## Abstract

Coral reefs support numerous marine organisms and are an important source of coastal protection from storms and floods, representing a major part of marine ecosystems. However coral reefs face increasing threats from pollution, ocean acidification, and sea temperature anomalies, making efficient protection and monitoring heavily urgent. Therefore, this study presents a novel machine-learning-based coral bleaching classification system based on a diverse global dataset with samples of healthy and bleached corals under varying environmental conditions, including deep seas, marshes, and coastal zones. We benchmarked and compared three state-of-the-art models: Residual Neural Network (ResNet), Vision Transformer (ViT), and Convolutional Neural Network (CNN). After comprehensive hyperparameter tuning, the CNN model achieved the highest accuracy of 88%, outperforming existing benchmarks. Our findings offer important insights into autonomous coral monitoring and present a comprehensive analysis of the most widely used computer vision models.

## Introduction

Coral reefs are marine ecosystems made up of colonies of invertebrates called corals. They are usually found in tropical and subtropical seas, offering protection, food, and feeding grounds for a large number of fish populations [1]. For example, the Great Barrier Reef, the largest living thing on the planet, is home to over 9,000 known species [2]. Furthermore, according to the United

Nations Environment Programme (UNEP), coral reefs cover less than 0.1 percent of the ocean but support over 25 percent of all marine creatures, hence playing a crucial role in ensuring biodiversity [3].

Coral bleaching occurs when corals expel the zooxanthellae algae due to environmental stress, causing the corals to turn white. Bleached corals are susceptible to disease and suffer reproductive issues that would often result in death [4, 5, 6]. These mass bleaching events are heavily disruptive to the marine ecosystem.

Furthermore, coral reefs are essential industries such as tourism and fishing, while also offering protection against dangerous storms and waves. Recently, coral bleaching events have been increasing at an exponential rate [7]. Current studies have shown that over 50 percent of the corals in the Great Barrier Reef experienced severe bleaching from 2016 to 2017 [8]. Furthermore, accounts have shown that the occurrence of bleaching is likely to increase with global warming and ocean acidification.

These phenomena are caused by the increase in sea temperatures, acidification of the oceans, pollution, and solar radiation. All of these factors are exacerbated by human activities, like increasing carbon and waste production [9, 10].

To address this issue, local governments have established Marine Protected Areas (MPAs), initiated coral restoration programs, and made commitments to reduce carbon emission [11, 12, 13, 14]. Specifically, in 2020, UNEP reported that 7.8% of global marine spaces were protected as MPAs, offering some protection to coral reefs from human intervention [15]. Furthermore, the Great Barrier Reef Foundation in 2015 launched a project designed to use underwater coral nurseries to transfer certain coral species [16]. The COP26 climate summit held in 2021 continued to push for urgent action in curbing emissions to prevent further loss of coral reefs [17]. Nations like the United States of America, Sweden, and the United Kingdom have also been pursuing and promoting the adoption of more renewable energy sources.

As machine learning object detection continues to advance, researchers and biologists have been using abundant data to use deep learning models to accurately identify and classify coral conditions. This paper is based on existing best practices literature, providing a robust and comprehensive analysis of new deep learning models and techniques. The remainder of this paper is organized as follows: Section 2 provides an extensive review of existing works. Section 3 describes data preprocessing, evaluation metrics, and model architectures. Section 4 gives detailed accounts of the performances of our models. Finally, this paper concludes in Section 5.

## Related Work

In recent years, there has been increasing attention towards coral preservation and protection in response to the escalating climate crisis. This section provides an overview of some important literature guiding the field of computer vision in coral monitoring.

Parsons et al. propose a UAV-based remote sensing methodology that integrates hyperspectral and RGB imaging with machine learning to enhance coral reef monitoring, enabling efficient classification of coral types and bleaching levels [18]. Raphael et al. review the application of deep learning for coral and benthic image classification, highlighting its ability to automate coral species recognition and detect biodiversity decline [19]. Wang et al. introduce ML-Net, a multi-local perception network for classifying healthy and bleached corals, which enhances classification accuracy by focusing on local coral features through multi-branch and multi-scale fusion blocks. ML-Net outperforms ResNet and ConvNext with an ACC result of 86.3 [20]. Burns et al. review machine learning algorithms for monitoring coral reef benthic composition, emphasizing the need for methods that can generalize spatially and temporally to improve scalability. They highlight convolutional and recurrent neural networks as promising frameworks [21]. White et al. applied a supervised machine learning algorithm to Landsat 8 imagery for mapping and monitoring the Great Barrier Reef, achieving an accuracy of 90%. The study demonstrates that remote sensing techniques can be effectively combined with classification algorithms to improve accuracy [22]. Boonnam et al. developed predictive models for coral reef bleaching under climate change using supervised and unsupervised machine learning techniques. Their study found that the Support Vector Machine (SVM) model achieved the highest accuracy of 88.85% in classifying bleaching levels; they also identified seawater pH and sea surface temperature as key factors in coral bleaching [23]. Hao et al. developed a multilabel deep learning method for classifying coral reef conditions and associated stressors using over 20,000 high-resolution underwater images; their ensemble learning approach outperformed existing algorithms [24]. Ai et al. proposed a novel coral reef classification method that combines the radiative transfer model with deep learning (RTDL) to enhance classification accuracy. Using ICESat-2 and Sentinel-2 data, the RTDL model improved classification performance by 5% [25]. Bovolo et al. proposed a novel fuzzy-input fuzzy-output support vector machine (FSVM) for subpixel image classification, integrating fuzzy logic to model mixed pixels and classify abundances of different classes, showing promising results [26]. Ren et al. proposed a novel CNN-XGBoost model for image classification, combining

CNN for feature extraction with eXtreme Gradient Boosting (XGBoost) for recognition, enhancing classification accuracy [27].

Despite the existing literature in the field of computer vision in coral classification, there are many research gaps that need to be addressed. Specifically, most previous studies struggled with accuracy rates above 80 percent due to poor model tuning or data augmentation. Furthermore, many existing studies focus on classifying corals in a niche location, often limiting the generalizability of models as they are not exposed to corals under varying environmental conditions. Additionally, most studies lack comparisons across models, offering a less comprehensive evaluation of model performances and uniqueness. Therefore, this study offers a more robust framework and detailed analysis to explore the aforementioned gaps.

## Methodology

### Objectives

This paper builds upon the existing literature to present a comprehensive evaluation and comparison of three state-of-the-art models—ResNet, ViT, and CNN—in the field of coral bleaching classification. Furthermore, this study introduces a robust framework for accurate coral classification in multi-condition underwater image datasets without requiring significant computational power. The key contributions can be summarized as follows:
- Presenting a comprehensive analysis of the accuracy of ResNet, ViT, and CNN.
- Demonstrating a robust framework for accurate classification in multi-condition coral imagery with minimal computational power requirements.
- Offering a detailed comparison between our results and existing benchmarks.

### Data

The dataset for this study was collected from Flickr using the Flickr API. It consists of a total of 923 labeled multi-condition underwater images, 438 of which are healthy and 485 of which are bleached, indicating minimal data imbalance. This dataset consists of a wide variety of coral species under different conditions, presenting a greater challenge to our deep learning models. Figure 1 presents sample images of bleached and healthy corals. The data distribution can be seen in Figure 2.

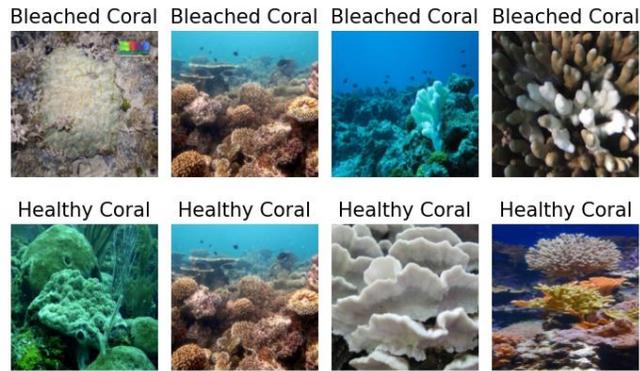

Fig. 1. Representative Sample Images from Both Classes in the Dataset.

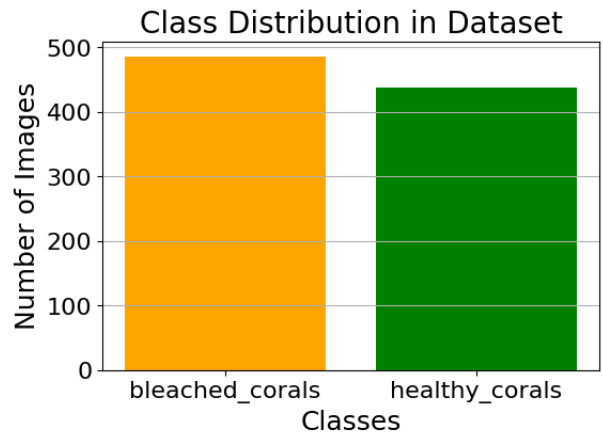

Fig. 2. Data Distribution of Healthy and Bleached Corals.

Furthermore, the images were resized to a maximum of 300 pixels for both height and width for more effective model training. After preprocessing, we split the dataset into three distinct sets: the training set, validation set, and testing set, based on a 70%, 15%, and 15% split, respectively. The data split distribution can be seen in Figure 3.

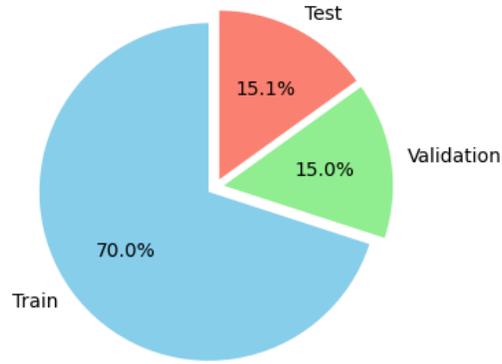

Fig. 3. Data Distribution Across Training, Validation, and Testing Sets.

**Evaluation Metrics**

To thoroughly assess our models, we utilized four common evaluation measures, namely precision, recall, F1 score, and accuracy. Precision measures how many of the predicted positive cases were truly correct. Higher precision indicates fewer false positives. Mathematical representation is as follows

$$\text{Precision} = \frac{TP}{TP + FP} \quad (1)$$

$TP$ represents True Positives and $FP$ represents False Positives. Recall measures how many actual bleaching cases were correctly identified. High recall means fewer false negatives. The mathematical notation is shown below.

$$\text{Recall} = \frac{TP}{TP + FN} \quad (2)$$

$FN$ represents False Negatives. F1-score balances precision and recall by including both false positives and false negatives. It includes values ranging from 0 to 1, where 1 indicates perfect precision and recall. The mathematical notation is shown below:

$$\text{F1-Score} = 2 \cdot \frac{Precision \cdot Recall}{Precision + Recall} \quad (3)$$

Accuracy determines the number of instances correctly predicted out of all the instances in the dataset to compute the overall accuracy. The mathematical formula is presented below:

$$\text{Accuracy} = \frac{Correct\ Predictions}{Total\ Predictions} = \frac{TP+TN}{Total\ Samples} \quad (4)$$

$TN$ represents True Negatives.

**Classification Models**

In this study, we utilized three powerful deep learning models to identify and classify coral bleaching. This section offers detailed descriptions of their architectures and characteristics, comparing their strength and weaknesses.

*Residual Neural Network (ResNet)*

ResNet is a deep learning neural network that is based on residual connections, making it robust to the vanishing gradient problem. The residual neural networks utilize skip connections, which allow the gradient to pass through numerous layers for training deeper networks. This network learns the residual rather than optimizing a transformation $y = F(x)$.

$$y = x + F(x) \quad (5)$$

$x$ is the input and $F(x)$ is the learned transformation. When the gradient updates are transmitted during backpropagation, we represent it as follows:

$$\frac{\partial y}{\partial x} = 1 + \frac{\partial F(x)}{\partial x} \quad (6)$$

As the gradient is preserved, residual neural networks can be used to train very deep networks without performance degradation. However, because of the depth of the architecture, it is memory intensive and can suffer from overfitting.

*Vision Transformer* **(ViT)**

Unlike traditional convolutional architectures, ViT processes an image as a sequence of patches and applies a self-attention mechanism to capture long-range dependencies. An input image $I$ is divided into non-overlapping patches, which are flattened and projected into a lower-dimensional space:

$$z_0 = [x_1 E;\ x_2 E;\ \dots;\ x_N E] + E_{pos} \quad (7)$$

$x_i$ represents the i-th image patch, $E$ is the learned embedding matrix, and $E_{pos}$ encodes positional information. The model then applies multi-head self-attention, defined as:

$$Attention(Q, K, V) = softmax\left(\frac{QK^T}{\sqrt{d_k}}\right)V \qquad (8)$$

$Q$, $K$, and $V$ represent the query, key, and value matrices, respectively, and $d_k$ is the dimensionality of the key vectors. The final classification is performed using a fully connected layer, which maps the learned class token $h_{CLS}$ to the output space:

$$y = W_o h_{CLS} + b_o \qquad (9)$$

$W_o$ and $b_o$ are the learned weight and bias parameters. Therefore, the self-attention mechanism allows ViT to capture global relationships between patches, unlike CNNs, which focus more on local features. However, ViT requires large datasets for training because it lacks built-in inductive biases of convolutions.

### *Convolutional Neural Network (CNN)*

CNNs process visual data using layers of convolutional filters that extract hierarchical features. The fundamental operation in CNNs is the convolution, where an image $I$ is processed using a kernel $K$:

$$(I * K)(x, y) = \sum_{i=-m}^{m} \sum_{j=-n}^{n} I(x+i, y+j) K(i,j) \qquad (10)$$

This operation captures spatial patterns such as edges and textures, enabling feature extraction at different levels. A non-linearity, typically a ReLU activation function, is applied to introduce element-wise transformations:

$$f(x) = \max(0, x) \qquad (11)$$

Pooling layers further reduce spatial dimensions while retaining essential features. A common approach is max pooling, which selects the highest value in a given receptive field:

$$p_{ij} = \max_{(m,n) \in R_{ij}} x_{mn} \qquad (12)$$

$R_{ij}$ represents the pooling region. The extracted features are then flattened and passed through fully connected layers for classification:

$$y = Wx + b \qquad (13)$$

$W$ and $b$ are learnable parameters. CNNs have demonstrated remarkable success in various vision tasks due to their ability to learn spatial hierarchies efficiently.

## Results and Discussion

After hyperparameter tunings and model training, the three models were evaluated against standard evaluation metrics to analyze their respective performances.

TABLE 1. PERFORMANCE EVALUATIONS

| *Model* | *Precision* | *Recall* | *F1-Score* | *Accuracy* |
|---|---|---|---|---|
| ResNet-50 | 0.86 | 0.86 | 0.86 | 0.86 |
| ViT | 0.64 | 0.64 | 0.64 | 0.64 |
| CNN | 0.89 | 0.88 | 0.88 | 0.88 |

As seen from Table 1, a standard CNN achieved superior accuracy, followed by ResNet-50 and ViT. Standard CNNs are capable of capturing local spatial hierarchies and dependencies, making them effective for many computer vision tasks. ResNet-50, on the other hand, while deeper, relies on residual connections, which may not be as beneficial depending on the dataset. ViT uses self-attention mechanisms and may struggle to understand local features without extensive tuning.

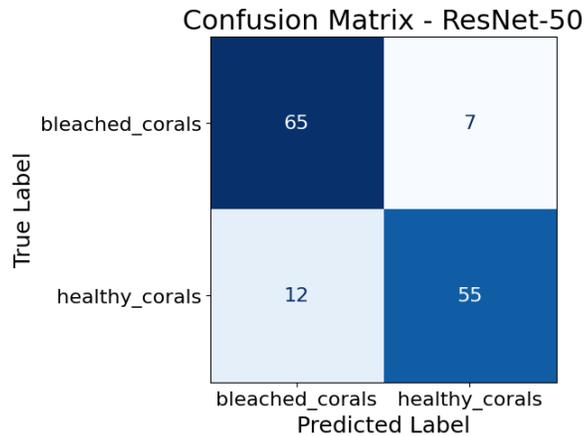

Fig. 4. Confusion Matrix for ResNet-50 Model.

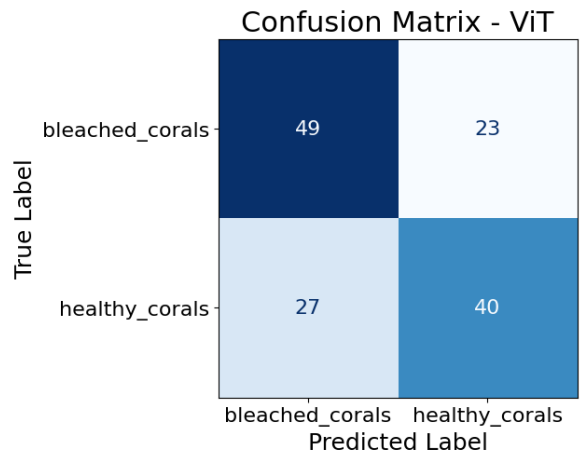

Fig. 5. Confusion Matrix for ViT Model.

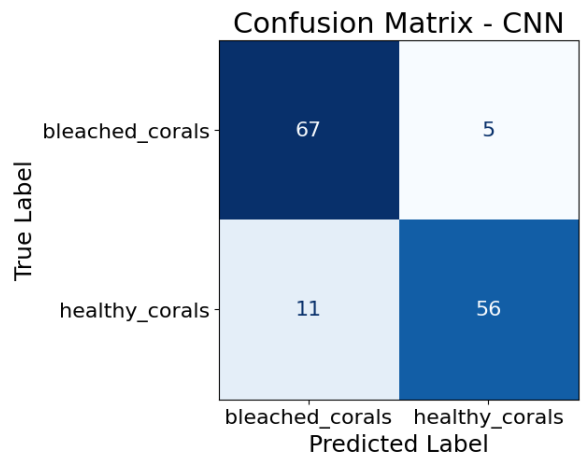

Fig. 6. Confusion Matrix for CNN Model.

As seen from Figures 4, 5, and 6, all three models are more capable of identifying bleached corals compared to healthy corals. Furthermore, it is evident that the performance of ViT significantly lags behind ResNet-50 and CNN.

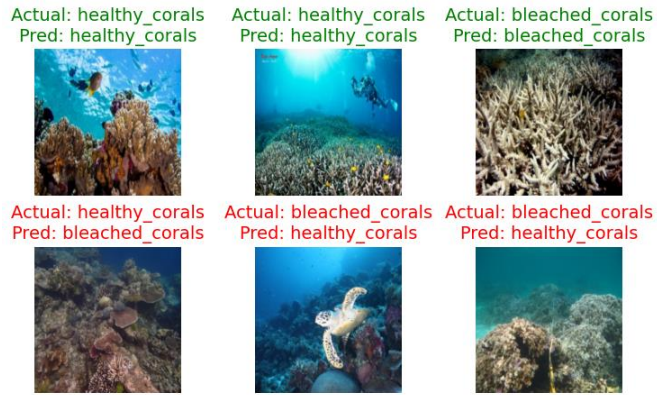

Fig. 7. Sample Predictions from ResNet-50 Model.

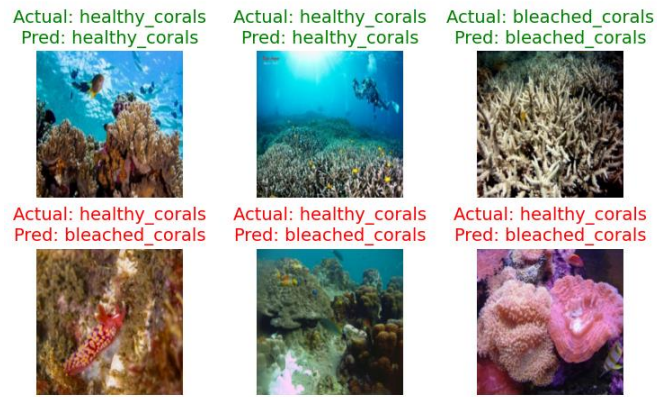

Fig. 8. Sample Predictions from ViT Model.

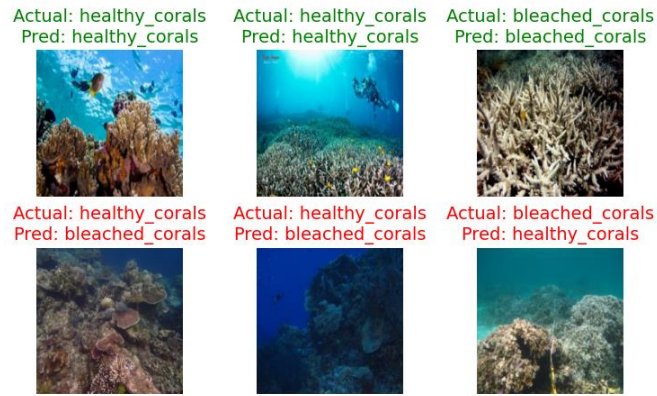

Fig. 9. Sample Predictions from CNN Model.

Figures 7, 8, and 9 show sample predictions from the models. Specifically, we can see that the accurately classified images are often much brighter and clearer, with the coral located in an easily detectable location. On the other hand, incorrectly classified images often feature unique corals with distinct colors or a hazy image.

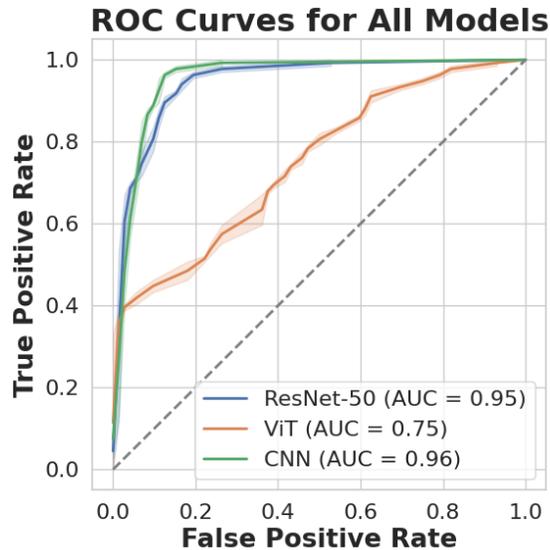

Fig. 10. Receiver Operating Characteristic (ROC) Curve.

Figure 10 features the ROC curve for the three models. An ROC graph visualizes a model's ability to distinguish between classes and find a balance between sensitivity and false positives. The Area Under the Curve (AUC) value provides the overall performance. The greater the AUC value, the better the model. As we can see, CNN achieved the highest AUC of 0.96, closely followed by ResNet with an AUC of 0.95, and lastly, ViT with an AUC of 0.75.

## Conclusion

Building upon existing literature and state-of-the-art models and techniques, this study presents a robust and comprehensive framework for accurately classifying coral bleaching with multi-condition underwater images. Specifically, we employed three deep learning computer vision models—ResNet-50, ViT, and CNN—to classify bleached and healthy corals. For evaluation, we utilized four standard metrics: precision, recall, F1-score, and accuracy for comprehensiveness. Furthermore, we analyzed our results exhaustively with detailed graphs and matrices. Our top model achieved an accuracy of 88%, demonstrating superior performance compared to previous studies. Moreover, our proposed framework is lightweight and flexible, offering invaluable insights for researchers and biologists alike.